%% file: root.tex
\documentclass[letterpaper, 10 pt, conference]{ieeeconf}  

\IEEEoverridecommandlockouts                              

\usepackage{graphics} 
\usepackage{amsmath} 
\usepackage{amssymb}  
\usepackage{cite}
\usepackage{listings}
\usepackage{xcolor}
\usepackage{graphicx}
\usepackage{subcaption}
\usepackage{hyperref} 
\usepackage{float}
\usepackage{bbm}
\usepackage{soul}
\usepackage{capt-of} 
\usepackage[most]{tcolorbox} 
\tcbuselibrary{listings,raster}

\newtcblisting{WideCode}{
  listing only,
  enhanced,
  breakable,
  colback=gray!3,
  colframe=black!25,
  boxrule=0.4pt,
  arc=2mm,
  left=1mm,right=1mm,top=1mm,bottom=1mm,
  listing options={
    language=Python,
    numbers=none,
    basicstyle=\ttfamily\footnotesize,
    columns=fullflexible,
    keepspaces=true,
    showstringspaces=false,
    breaklines=true,
    tabsize=2
  }
}

\definecolor{codegreen}{rgb}{0,0.6,0}
\definecolor{codegray}{rgb}{0.5,0.5,0.5}
\definecolor{codepurple}{rgb}{0.58,0,0.82}
\definecolor{backcolour}{rgb}{0.95,0.95,0.92}

\usepackage[scaled=0.9]{sourcecodepro} 
\lstdefinestyle{mystyle}{
    backgroundcolor=\color{backcolour},   
    commentstyle=\color{codegreen},
    keywordstyle=\color{magenta},
    numberstyle=\tiny\color{codegray},
    stringstyle=\color{codepurple},
    basicstyle=\ttfamily\footnotesize,
    xleftmargin=1.6em,       
    framexleftmargin=1.6em,  
    breakatwhitespace=false,         
    breaklines=true,                 
    captionpos=b,                    
    keepspaces=true,                 
    numbers=left,                    
    numbersep=5pt,                  
    showspaces=false,                
    showstringspaces=false,
    showtabs=false,                  
    tabsize=2,
    language=Python
}

\title{\LARGE \bf
SRMP: Search-Based Robot Motion Planning Library
}

\author{\authorblockN{\textbf{Itamar Mishani$^{*}$, Yorai Shaoul$^{*}$, Ramkumar Natarajan$^{*\dag}$\thanks{$\dag$ Currently at Amazon Robotics. This work was done outside of Amazon.}, Jiaoyang Li, and Maxim Likhachev}}
\authorblockA{The Robotics Institute, School of Computer Science, Carnegie Mellon University \\
$*$ Equal Contribution \\
\texttt{\{imishani, yshaoul, rnataraj, jiaoyanl, maxim\}@cs.cmu.edu}}}

\newcommand{\conf}[2]{\mathbf{q}^{#1}_{#2}}

\begin{document}

\maketitle


\input{sections/abstract}
\input{sections/introduction}

\input{figures/code_example}
\input{sections/srmp}

\input{sections/experiments}

\input{sections/background} 
\input{sections/conclusion}
\input{sections/acknowledge}

\bibliographystyle{IEEEtran}
\bibliography{root}
\end{document}

%% file: sections/abstract.tex
\begin{abstract}
Motion planning is a critical component in any robotic system.
Over the years, powerful tools like the Open Motion Planning Library (OMPL) have been developed, offering numerous motion planning algorithms. However, existing frameworks often struggle to deliver the level of predictability and repeatability demanded by high-stakes applications---ranging from ensuring safety in industrial environments to the creation of high-quality motion datasets for robot learning. Complementing existing tools, we introduce SRMP (Search-based Robot Motion Planning), a new software framework tailored for robotic manipulation. SRMP distinguishes itself by generating consistent and reliable trajectories, and is the first software tool to offer motion planning algorithms for multi-robot manipulation tasks. SRMP easily integrates with major simulators, including MuJoCo, Sapien, Genesis, and PyBullet via a Python and C++ API. SRMP includes a dedicated MoveIt! plugin that enables immediate deployment on robot hardware and seamless integration with existing pipelines. Through extensive evaluations, we demonstrate in this paper that SRMP not only meets the rigorous demands of industrial and safety-critical applications but also sets a new standard for consistency in motion planning across diverse robotic systems. Visit \href{https://srmp.readthedocs.io}{\texttt{srmp.readthedocs.io}} for SRMP documentation and tutorials.
\end{abstract}

%% file: sections/introduction.tex
\section{INTRODUCTION}

The field of robotics is rapidly becoming more accessible to a wide range of practitioners. As tools for experimenting with robots have become easier to use, programmers with minimal robotics experience can now develop robotic applications in just a few lines of code. For instance, modern physics simulators such as MuJoCo \cite{todorov2012mujoco}, Sapien \cite{xiang2020sapien}, and Genesis \cite{Genesis} enable users to simulate robotic systems effortlessly using Python, lowering the barrier to entry. This increased accessibility has led to a growing demand for motion planning libraries that are powerful, reliable, and integrable with modern robotic frameworks and simulators.

\input{figures/fig_teaser}

Many great tools for motion planning have been developed over the past two decades. The Open Motion Planning Library (OMPL) \cite{ompl}, for instance, is widely used for its extensive collection of sampling-based algorithms, efficient implementations, and availability in popular robotics tools \cite{moveit, mplib}. However, OMPL, like many motion planning libraries that use sampling-based algorithms, has certain limitations that hinder its applicability for some real-world robotics use cases. Most notably, the solutions generated by sampling-based planners often exhibit significant variability -- a negative property when consistency and interpretability of motions are important.

\input{figures/fig_single_agent_consistency}

Variability in generated motions can be particularly problematic in industrial applications and modern robot learning systems. In industrial settings, robots and humans often share a workspace, and for safe collaboration, human workers must be able to anticipate robot motions. This can be achieved through consistent and repeatable motion planning algorithms. Similar properties are also beneficial for robot learning, especially with recent approaches that increasingly rely on motion planners \cite{janner2022diffuser, mpd, diffusionpolicy}. For example, MPD \cite{mpd} found success in relying on motion planners for collecting demonstration data for trajectory learning, and PerAct \cite{shridhar2022peract} reduced the learning problem to sequences of end-effector key-poses and assumed access to reliable motion planners for moving between key-poses during policy execution. Existing tools, which mainly focus on finding \textit{some} collision-free trajectory, may lead to unpredictable, potentially dangerous motions, inconsistent datasets, and variable execution results. 

To address these challenges, we introduce \textit{Search-based Robot Motion Planning} (SRMP): a software framework designed for consistent, safe, reliable and repeatable motion generation (Fig. \ref{fig:teaser}). SRMP achieves this by leveraging state-of-the-art graph search algorithms with strong theoretical guarantees of completeness and bounded sub-optimality, which translate to short and repeatable motions. In practice, SRMP integrates seamlessly with widely used tools, including MoveIt!, various physics simulators, and numerous real-world robotic platforms. Its ease of use and strong algorithmic foundations make SRMP an effective motion planning framework for modern robotics applications, including data collection, industrial automation, and research. Our contributions are as follows:
\begin{enumerate}
    \item \textbf{SRMP Software Framework}: a new motion planning library for robotic manipulation that delivers reliable, repeatable, and interpretable plans through powerful search-based algorithms. SRMP directly overcomes key limitation of existing libraries by providing algorithms that are unavailable elsewhere.
    \item \textbf{Support for Scale}: SRMP is the first motion planning library to support both single-robot and multi-robot manipulation planning.
    \item \textbf{Seamless Integration}: SRMP integrates with modern robotics tools, including MoveIt!, physics simulators (MuJoCo, Sapien, Genesis, etc.), and real hardware.
    \item \textbf{Multi-Language Support:} SRMP's C++ optimized backend and Python APIs enable accessibility for both researchers and industry practitioners.
\end{enumerate}

\noindent SRMP bridges the gap between academic research and practical robotics applications and provides a robust, scalable solution for motion planning in simulation and real-world deployments. We begin by introducing SRMP (Sec. \ref{sec:srmp}), proceed with a discussion of experimental results (Sec. \ref{sec:experiments}), and close with relevant background (Sec. \ref{sec:background}).

%% file: figures/fig_teaser.tex
\begin{figure}[!t]
    \centering
    \includegraphics[width=0.99\linewidth]{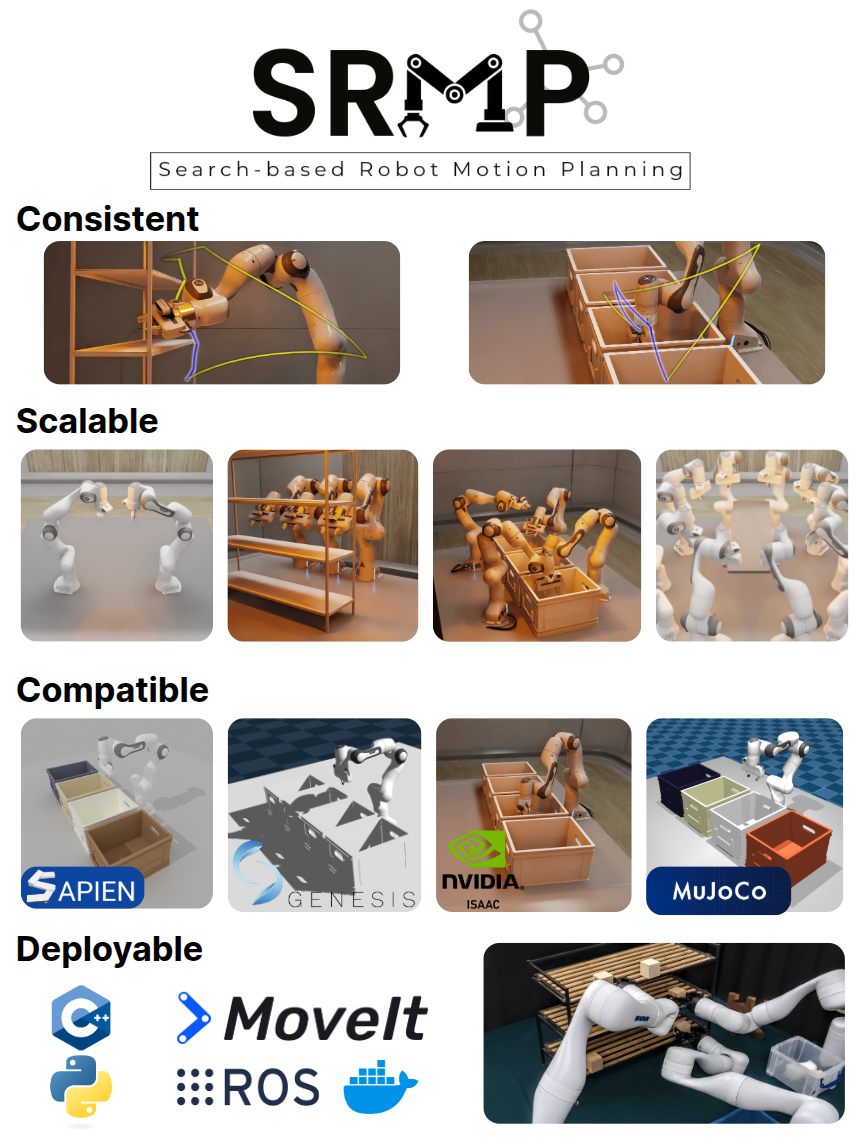}
    \caption{SRMP is a robot motion planning software that efficiently generates consistent motion plans for one or many robots. SRMP supports modern simulators, is available through C++ and Python APIs, and integrates with MoveIt! via a plugin.}
    \label{fig:teaser}
    \vspace{-0.5cm}
\end{figure}

%% file: figures/fig_single_agent_consistency.tex
\begin{figure*}[t]
    \centering
    \begin{subfigure}[b]{0.18\textwidth}
        \centering
        \includegraphics[width=\textwidth]{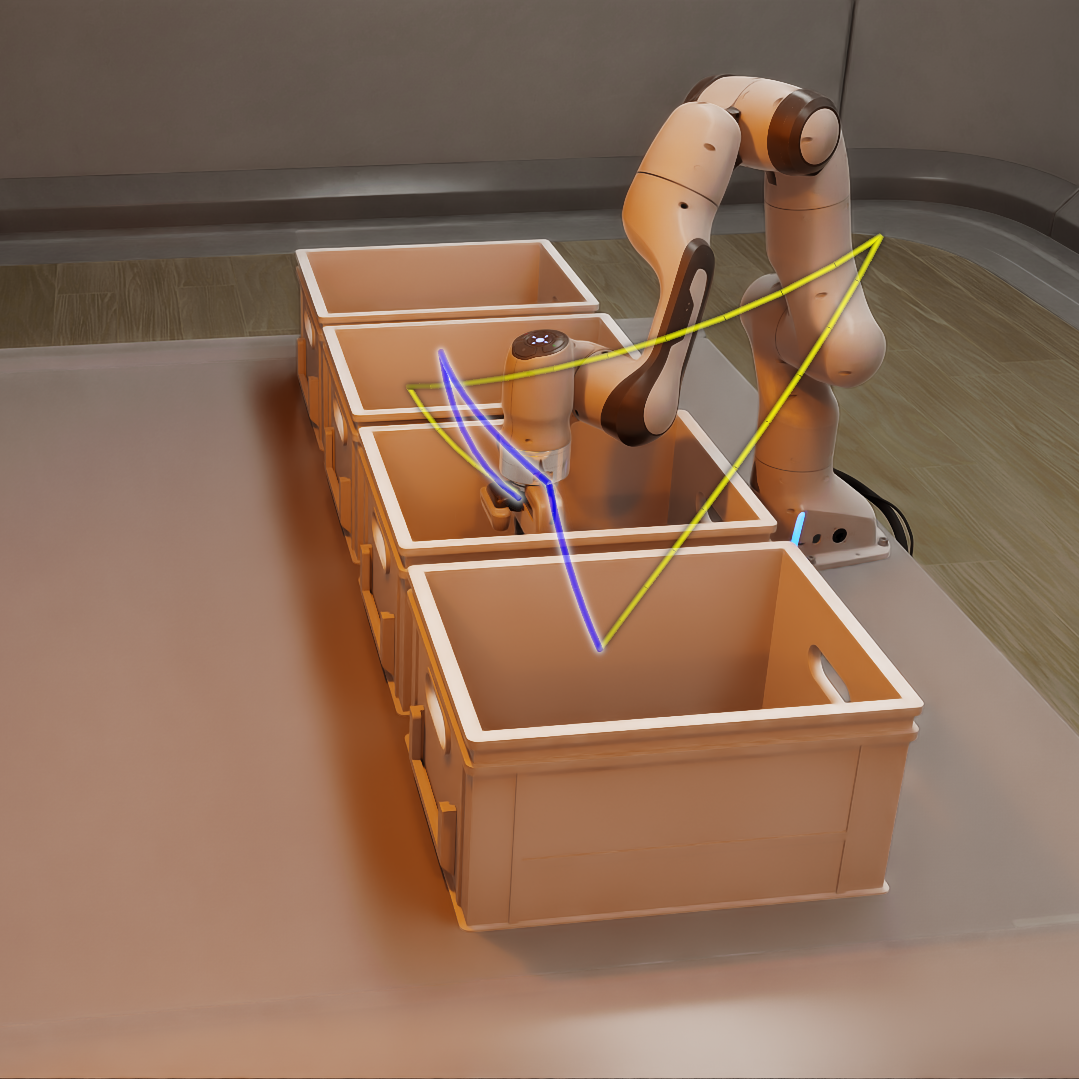}
    \end{subfigure}
    \centering
    \begin{subfigure}[b]{0.24\textwidth}
        \centering
        \includegraphics[width=\textwidth]{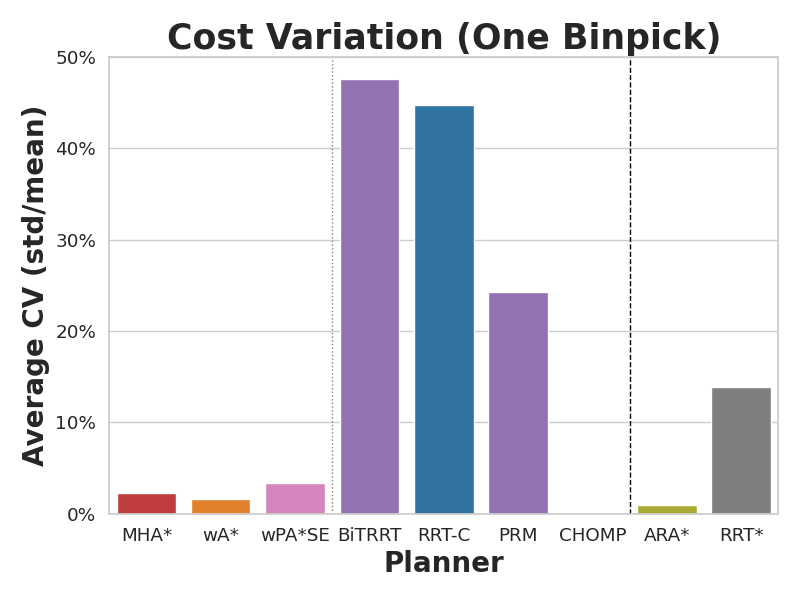}
    \end{subfigure}
    \hfill
    \begin{subfigure}[b]{0.3\textwidth}
        \centering
        \includegraphics[width=\textwidth]{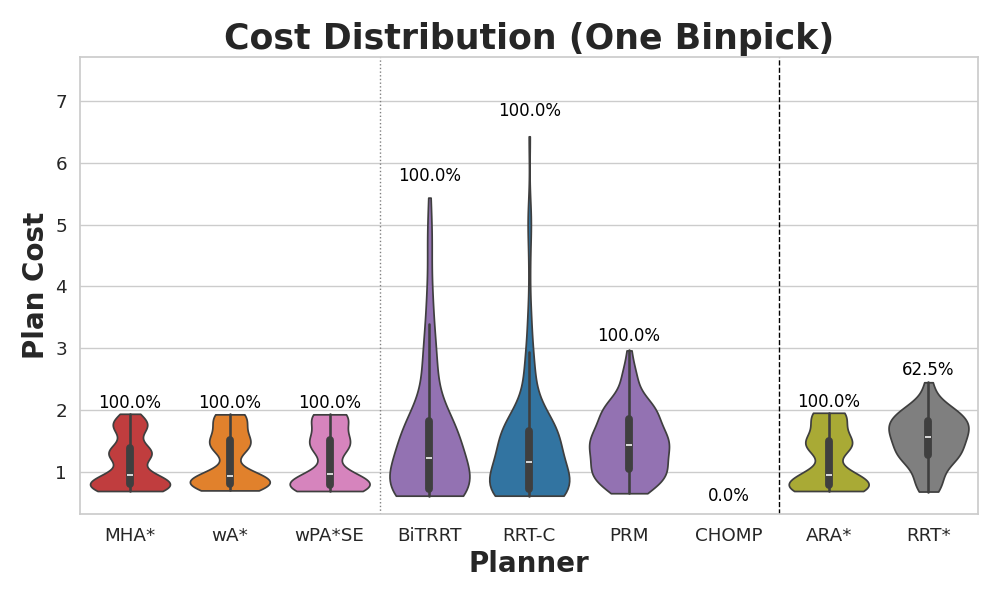}
    \end{subfigure}
    \hfill
    \begin{subfigure}[b]{0.24\textwidth}
        \centering
        \includegraphics[width=\textwidth]{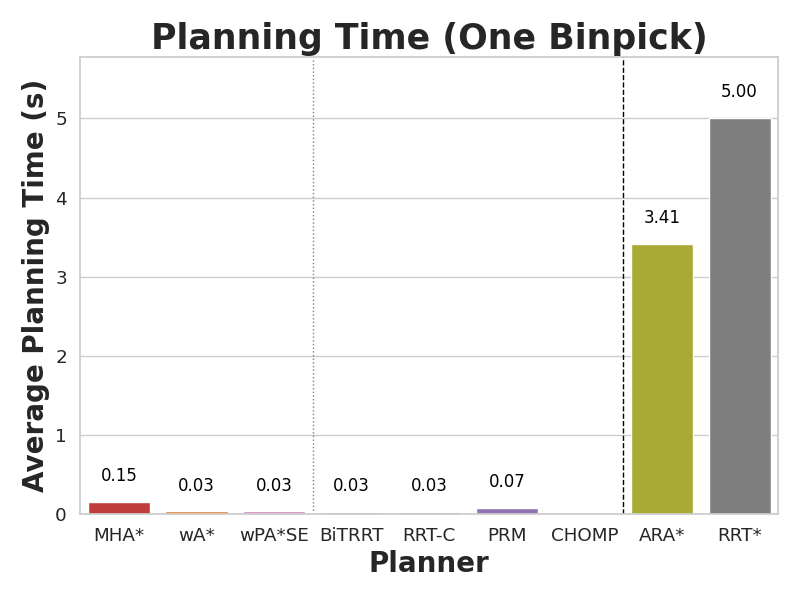}
    \end{subfigure}
    \centering
    \begin{subfigure}[b]{0.18\textwidth}
        \centering
        \includegraphics[width=\textwidth]{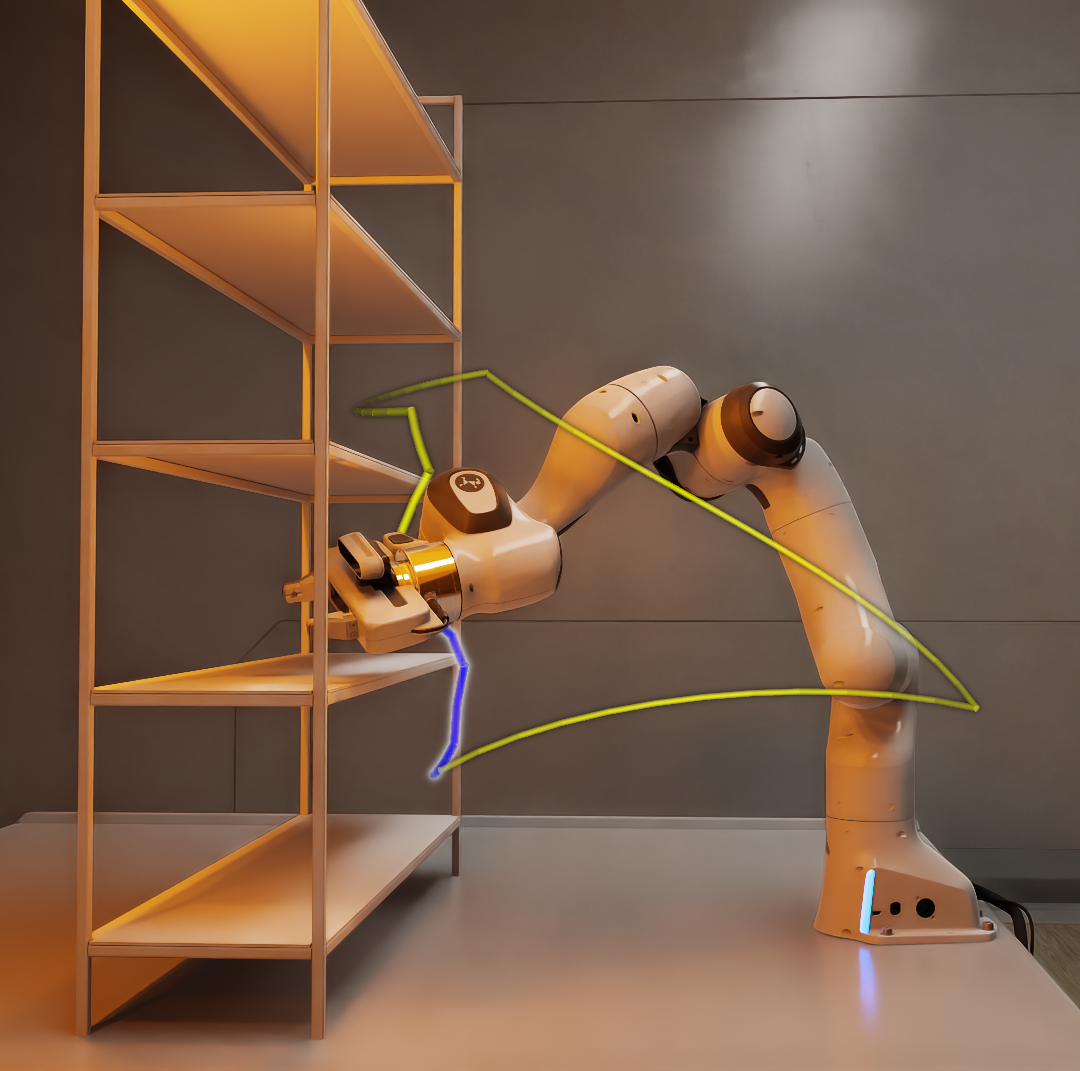}
    \end{subfigure}
    \centering
    \begin{subfigure}[b]{0.24\textwidth}
        \centering
        \includegraphics[width=\textwidth]{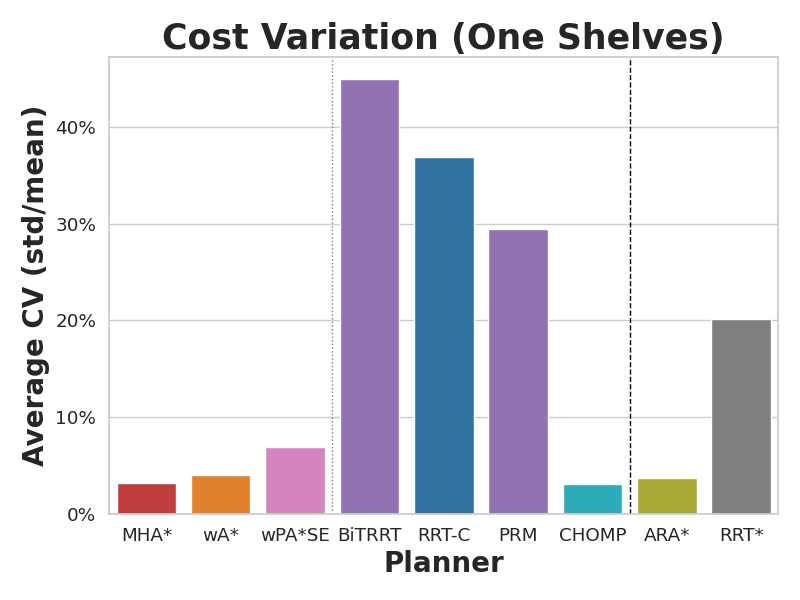}
    \end{subfigure}
    \hfill
    \begin{subfigure}[b]{0.3\textwidth}
        \centering
        \includegraphics[width=\textwidth]{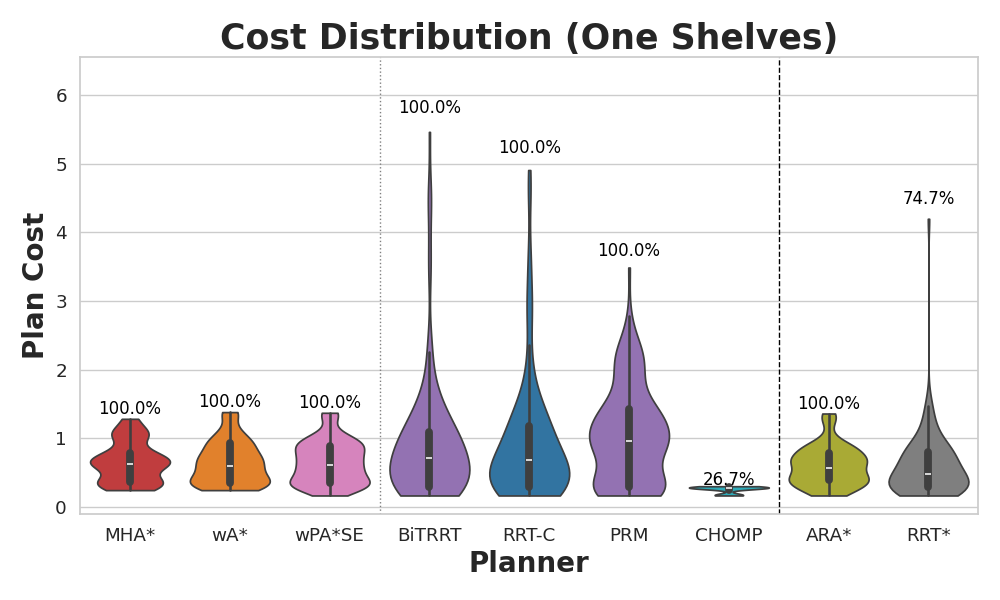}
    \end{subfigure}
    \hfill
    \begin{subfigure}[b]{0.24\textwidth}
        \centering
        \includegraphics[width=\textwidth]{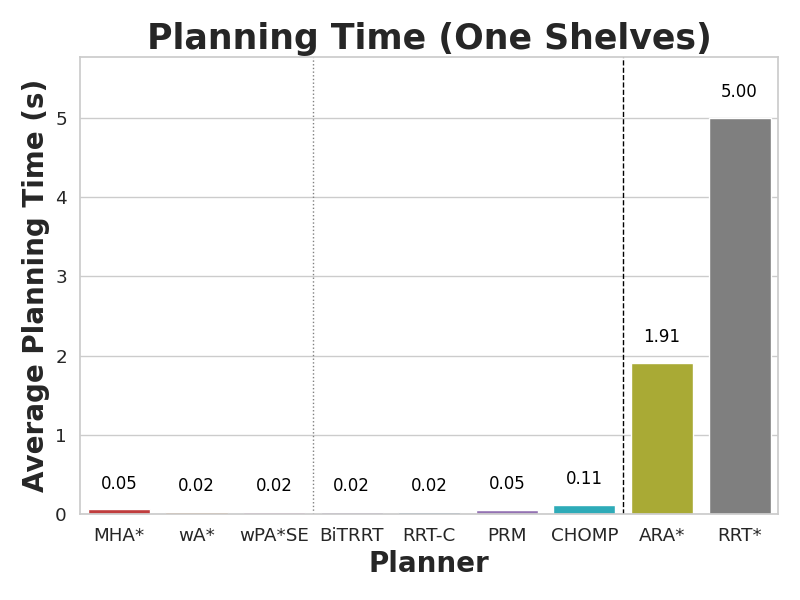}
    \end{subfigure}
    \caption{Single arm consistency evaluations. Top: Bin picking and placing. Bottom: Shelf-to-shelf movement. For each scenario, algorithms repeated each of 15 planning problems 10 times, every time with a slight variation in their goal configuration.  We compare intra-problem coefficient of variance (CV), overall average solution costs (annotated with success rates), and planning times. On the right of each figure, two \textit{anytime} algorithms -- ARA* and RRT* -- run until finding an optimal solution or reaching a five-second time limit. The left-side images illustrate our experimental scenarios, alongside an example of an end-effector trajectory generated using SRMP (blue) and one generated by OMPL (yellow).}
    \label{fig:exps_single}
    \vspace{-.4cm}
\end{figure*}

%% file: figures/code_example.tex
\begin{figure*}[t]
\centering
\begin{tcolorbox}[enhanced, breakable, width=\textwidth,
  colback=backcolour, colframe=black!25, boxrule=0.4pt, arc=2mm,
  left=1mm,right=1mm,top=1mm,bottom=1mm]

\begin{minipage}[t]{0.5\textwidth}
\begin{lstlisting}
import srmp
planner = srmp.PlannerInterface()
planner.add_articulation(urdf_path="panda.urdf", 
                    srdf_path="panda.srdf", 
                    name="panda", 
                    end_effector="panda_hand")
planner.add_box("box", size=np.array([1,2.1,1]), 
                pose=srmp.Pose(p=[0,0,0.4]))
planner.add_mesh("shrimp", path="shrimp.stl", 
                 pose=srmp.Pose(p=[2,0,1], 
                                q=[0,0,0,1]))
q_start = np.radians([0, -45, 0, -135, 0, 90, 45])
# Define goal constraint (e.g., gripper pose).
goal = srmp.GoalConstraint(srmp.GoalType.POSE,
       list(srmp.Pose([0.6, -0.0, 0.5])))
planner.make_planner(["panda"],
                     {"planner_id": "ARAstar",
                      "time_limit":"5"}) 
trajectory = planner.plan(["panda"], 
                           q_start, goal)
\end{lstlisting}
\end{minipage}
\hfill
\begin{minipage}[t]{0.50\textwidth}
\begin{lstlisting}[firstnumber=23]
# Add a second robot model to the world
planner.add_articulation(urdf_path="panda.urdf", 
                         srdf_path="panda.srdf", 
                         name="panda2", 
                         end_effector="panda_hand")
planner.set_base_pose(name="panda2", 
        pose=srmp.Pose(p=[0.5, 0, 0], 
                       q=[0,0,0,1]))
# Multi-agent goal constraint
goal_2 = srmp.GoalConstraint(
        srmp.GoalType.POSE,
        list(srmp.Pose([0.6, -0.5, 1.0])))
# Configure multi-agent planner
planner.make_planner(["panda","panda2"],
                     {"planner_id":"xECBS",
                      "time_limit":"5"})
start_qs = {"panda":  q_start,
            "panda2": q_start}
goals =    {"panda":  goal,
            "panda2": goal_2}
trajectory = planner.plan_multi(start_qs, goals)
\end{lstlisting}
\end{minipage}
\end{tcolorbox}
\caption{Planning for a single robot (left) and multiple robots (right) with the \texttt{srmp} Python API.}
\label{fig:code}
\vspace{-0.3cm}
\end{figure*}

%% file: sections/srmp.tex
\section{SRMP}
\label{sec:srmp}
SRMP is a search-based motion planning software providing a set of algorithms for solving motion planning problems for robotic manipulation. 
Built as a standalone library, SRMP can be integrated into existing pipelines in a plug-and-play manner, including MoveIt! (a widely used motion planning framework) as a plugin. 
SRMP is implemented in C++ for performance and provides Python bindings for ease of use.

\subsection{Usage}
SRMP can be used in two main ways: as a standalone library or as a MoveIt plugin. In this section, we describe how to use the library in Python, but the same functionality is available in C++ and will be documented accordingly.

\subsubsection{Standalone Library}
The primary interface for using the library is through the \texttt{PlannerInterface} class, which provides an abstraction for interacting with the planning world.
Using \texttt{PlannerInterface}, users can load a robot model into the planning world, add and remove objects -- including primitive shapes, meshes, and point clouds -- and configure the planner for a given start configuration and goal condition. The planner returns a trajectory consisting of a sequence of configurations, velocities, accelerations, and time stamps. SRMP supports planning for multiple manipulators, either independently or as a coordinated multi-agent system.

The code snippet in Fig. \ref{fig:code} illustrates the basic usage of SRMP in Python. The code loads a robot model, adds objects to the environment, defines a start configuration, and specifies a goal condition. The planner is then configured and used to compute a trajectory from the start configuration to the goal condition.
In the right column of the figure, an additional robot is added to the environment, and a trajectory is computed for both robots using multi-agent planning.
SRMP performs collision checking with COAL \cite{coalweb} and kinematic computations with Pinocchio \cite{carpentier2019pinocchio}.

\subsubsection{MoveIt Plugin}
SRMP can be used as a plugin for MoveIt! \cite{moveit}, a widely used motion planning framework. The plugin provides a bridge between MoveIt! and SRMP, allowing users to leverage the search-based motion planning capabilities of SRMP within the MoveIt! framework. 
The plugin is implemented as a MoveIt! planning adapter, recives planning requests from MoveIt! and forwards them to SRMP for processing.
The kinematics and collision checking modules of MoveIt! are used to provide the necessary information to SRMP. The plugin then uses the SRMP planner to compute a trajectory, which is returned to MoveIt! for execution.

To use SRMP as a MoveIt! plugin, users need to install the SRMP library and configure MoveIt! to use the SRMP planning adapter:

\begin{tcolorbox}[enhanced, breakable,
  colback=backcolour, colframe=black!25, boxrule=0.4pt, arc=2mm,
  left=1mm,right=1mm,top=1mm,bottom=1mm]
\begin{lstlisting}[language=Python, label=lst:srmp_moveit, numbers=none]
moveit::planning_interface::MoveGroupInterface move_group("panda_arm");
move_group.setPlanningPipelineId("srmp");
move_group.setPlannerId("wPASE");
move_group.setJointValueTarget(goal_state);
move_group.setPlanningTime(5.0);
move_group.move();
\end{lstlisting}
\end{tcolorbox}


%% file: sections/experiments.tex

\section{Experimental Analysis}
\label{sec:experiments}
To demonstrate the practical utility of SRMP, we conducted an experimental analysis of its performance on a variety of manipulation and multi-arm motion planning problems. We compare SRMP with the de-facto standard motion planning library, OMPL, and show that SRMP produces more consistent and repeatable solutions in comparable planning times while also scaling to multi-robot motion planning problems. In this section, we detail our experimental analysis and discuss its results. Later, in Sec. \ref{sec:background}, we formalize the motion planning problem and situate SRMP within the context of existing motion planning libraries and algorithms.
\vspace{-.1cm}

\input{figures/fig_multi_agent_consistency}

\subsection{Experimental Setup}
The aims of our analysis were first to assess the ability of SRMP to produce high-quality solutions to practical motion planning problems \textit{consistently}, and second to evaluate SRMP's ability to solve difficult motion planning problems. To assess consistency, we constructed four planning scenarios (two for single-robot, Fig. \ref{fig:exps_single}, and two for multi-robot, Fig. \ref{fig:exps_multi}) and generated random planning problems in each by choosing start and goal configurations from a collection of collision-free, ``task-relevant'' configurations (e.g., robots reaching into bins or between shelves).
Each planning problem was repeated multiple times (10 for single-robot problems and 5 for multi-robot problems) to assess the consistency of the algorithms. Each repetition was generated by perturbing the goal configuration slightly ($\pm 2^\circ$ in one joint), shedding light on the ability of the algorithms to find similar solutions to similar problems.

To understand how SRMP fares in general motion planning problems in the traditional sense, for the single-robot case, we evaluate our software using the Motion Benchmarker \cite{chamzas2022-motion-bench-maker} and report the results for a representative set of problems in Fig. \ref{fig:benchmark}.

\subsection{Evaluation Metrics}
The primary metrics we used throughout our analysis were \textit{cost} and \textit{consistency}. We also report on the \textit{planning time} for each algorithm and its \textit{success rate}.
In this analysis, we defined the cost of a path as the distance (in meters) traversed by its end-effector. In the multi-robot case, the cost was the sum of individual robot costs. This metric is useful for evaluating the efficiency of trajectories, and also a good indicator of how ``interpretable'' solutions are. That is, shorter end-effector motions are often perceived as more natural.

To assess the consistency of algorithms, we computed the average \textit{coefficient of variation} (CV) for each algorithm (in $\%$). Specifically, let $\{c_1, c_2, \ldots, c_n\}$ be the costs of the paths found by an algorithm across $n$ repetitions of a planning problem. The CV is defined as the ratio of the standard deviation to the mean, i.e., $\text{CV} = \frac{\sigma}{\mu}$, where $\sigma$ is the standard deviation of the costs and $\mu$ is the mean cost. A low CV indicates that the values are close to the mean, while a high CV indicates that the values are spread out over a large range. In the context of motion planning, 
a high CV indicates that the algorithm is producing paths of varying quality.

Finally, we measured the planning time of each algorithm in seconds and report the average time across all planning problems. We also report the success rate of each algorithm, which is the percentage of planning queries for which the algorithm was able to find a solution under the given time limit. We allowed algorithms to compute for $5$ seconds for single-robot problems and $20$ seconds for multi-robot. In multi-robot tests, we also report average \textit{effective runtime ratios}. For each planning problem, we found the shortest solution time $t_{\min{}}$ and scored each planner with $\frac{t}{t_{\min{}}}$ where $t$ is the time the planner spent planning.

\subsection{Evaluated Algorithms}
\textbf{Single-Robot Problems :} We compared SRMP's \textit{Multi-Heuristic A*} (\textit{MHA*}) \cite{aine2016mhastar}, weighted \textit{Parallel A* for Slow Expansions} (\textit{PA*SE}) \cite{pase} and \textit{Weighted-A*} (\textit{wA*}) \cite{a*} algorithms with OMPL's \textit{RRT Connect} (\textit{RRT-C}) \cite{RRT-Connect}, \textit{Bi-directional Transition-based RRT} (\textit{BiTRRT}) \cite{bitrrt}, \textit{BiEST} \cite{BiEST}, and \textit{PRM} \cite{PRM} algorithms, which are arguably the most popular algorithms for single-robot motion planning. We additionally included \textit{anytime} algorithms in our analysis -- algorithms that can recover shorter paths by running for longer periods of time. Here, we compare SRMP's \textit{Anytime Repairing A*} (\textit{ARA*}) \cite{anytime_dynamicA*} with OMPL's \textit{RRT*} \cite{RRT*}. We have also evaluated SRMP against \textit{CHOMP} \cite{ratliff2009chomp}, an optimization-based method available within the MoveIt! pipeline.

\textbf{Multi-Robot Problems:} We compared SRMP's \textit{xECBS} \cite{shaoulmishani2024xcbs} with OMPL's \textit{BiTRRT}, \textit{PRM}, and \textit{RRT-Connect} algorithms, which, despite not being designed for multi-robot problems, are commonly used in practice. In these tests, we focused on ``practical'' numbers of robots (up to 4), but note that SRMP can scale to larger numbers of robots \cite{shaoulmishani2024xcbs}.

\subsection{Results}
The first part of our experimental analysis (Fig. \ref{fig:exps_single} and Fig. \ref{fig:exps_multi}), which focused on solution quality and consistency, revealed that SRMP drastically outperformed OMPL in terms of cost and consistency. In single-robot problems, outlined in Fig. \ref{fig:exps_single}, SRMP's algorithms were significantly more consistent than those of OMPL. The solution costs from SRMP were almost always lower, and were obtained with comparable planning times. Among the anytime algorithms, SRMP's \textit{ARA*} algorithm recovered the \textit{optimal} path in some of the test cases, terminating its search early. This comes in contrast to OMPL's \textit{RRT*} algorithm that always ran until reaching the runtime limit. 

In multi-robot problems, outlined in Fig. \ref{fig:exps_multi}, similar but more pronounced trends were observed. As shown in Fig. \ref{fig:exps_multi}, SRMP's \textit{xECBS} algorithm produced paths that were shorter than those produced by OMPL's algorithms. Furthermore, SRMP's \textit{xECBS} algorithm was able to find solutions for almost all planning problems, while OMPL's algorithms failed to find solutions for at least $20\%$ of the problems. At an average planning time of 1.5 seconds for ``Binpick'' and 2.41 seconds for ``Shelves'', \textit{xECBS}'s planning time is longer than that of OMPL's algorithms (average 0.1 and 0.14 seconds, respectively), in part due to solving more difficult problems than the other algorithms. \cite{shaoulmishani2024xcbs} compares these algorithms in depth and shows more drastic differences in more difficult problems.

In our additional stress tests, outlined in Fig. \ref{fig:benchmark}, we integrated the Motion Benchmarker \cite{chamzas2022-motion-bench-maker} with SRMP. We note that the aforementioned trend generalized across domains and embodiments. The solution costs of motion plans found by SRMP were consistently lower than those of OMPL while maintaining only slightly slower average planning times and a perfect success rate.
The heuristic functions used in these tests for wA*, ARA*, wPA*SE, and MHA* were the Euclidean distance in the joint space and the backward breadth-first search (BFS) from the goal in the task-space. 

\input{figures/fig_benchmark}

%% file: figures/fig_multi_agent_consistency.tex
\begin{figure*}[!t]
    \centering
    \begin{subfigure}[b]{0.18\textwidth}
        \centering
        \includegraphics[width=\textwidth]{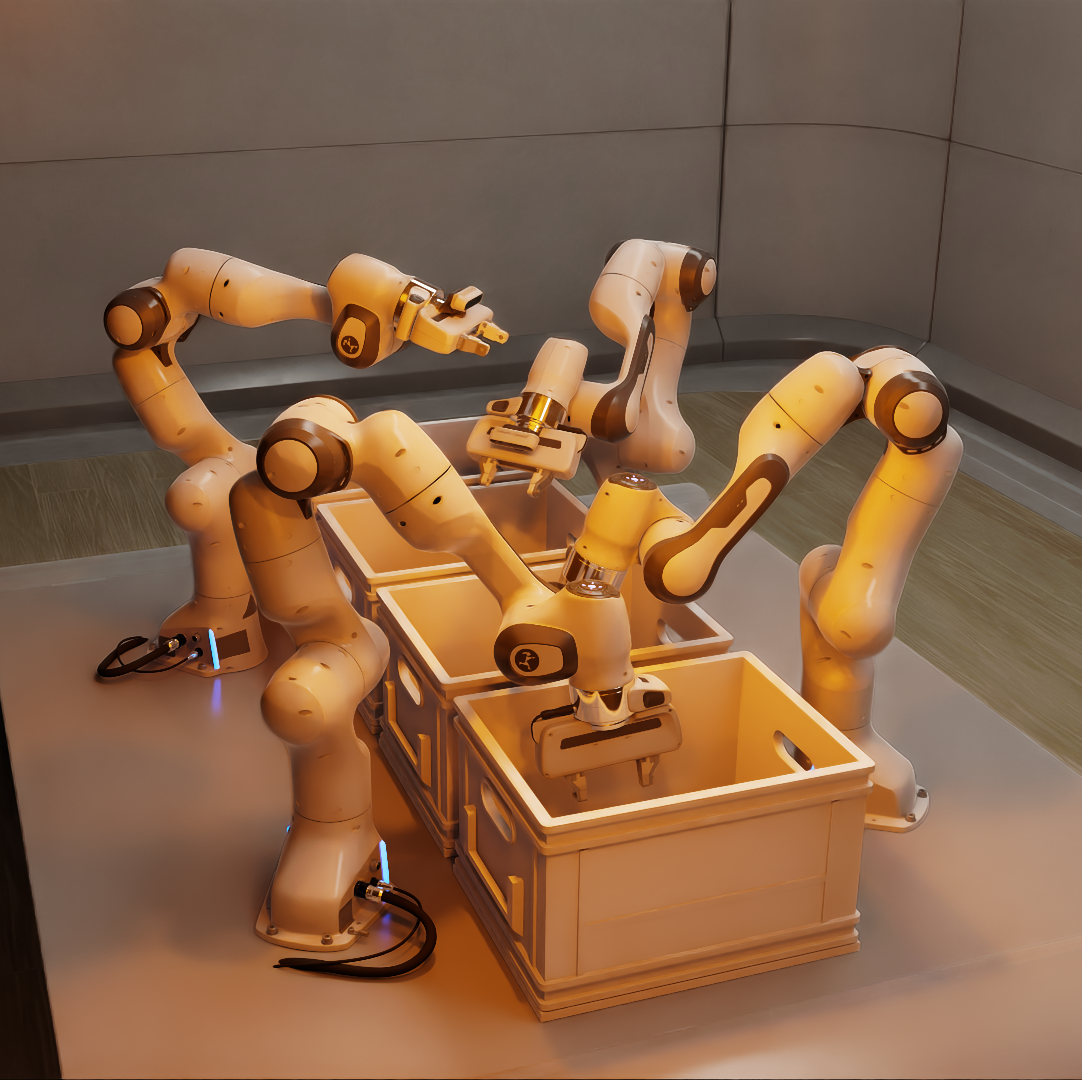}
    \end{subfigure}
    \centering
    \begin{subfigure}[b]{0.22\textwidth}
        \centering
        \includegraphics[width=\textwidth]{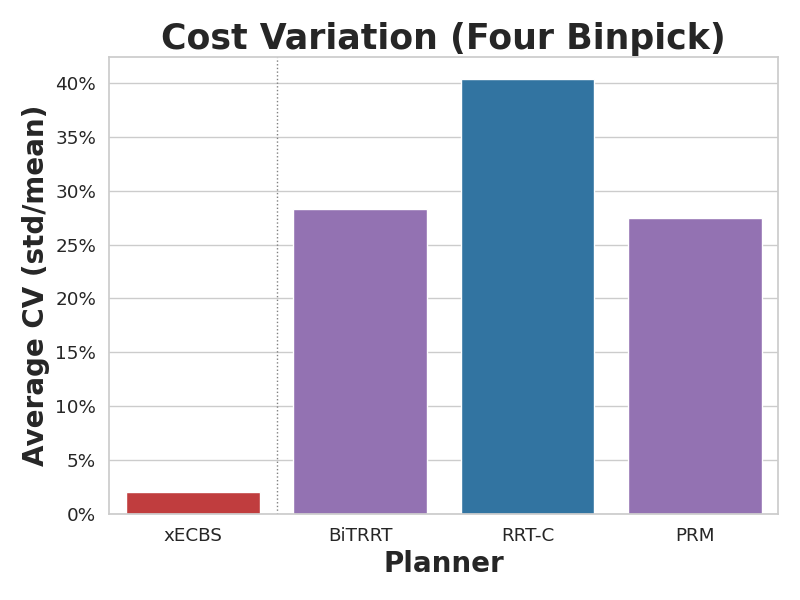}
    \end{subfigure}
    \hfill
    \begin{subfigure}[b]{0.28\textwidth}
        \centering
        \includegraphics[width=\textwidth]{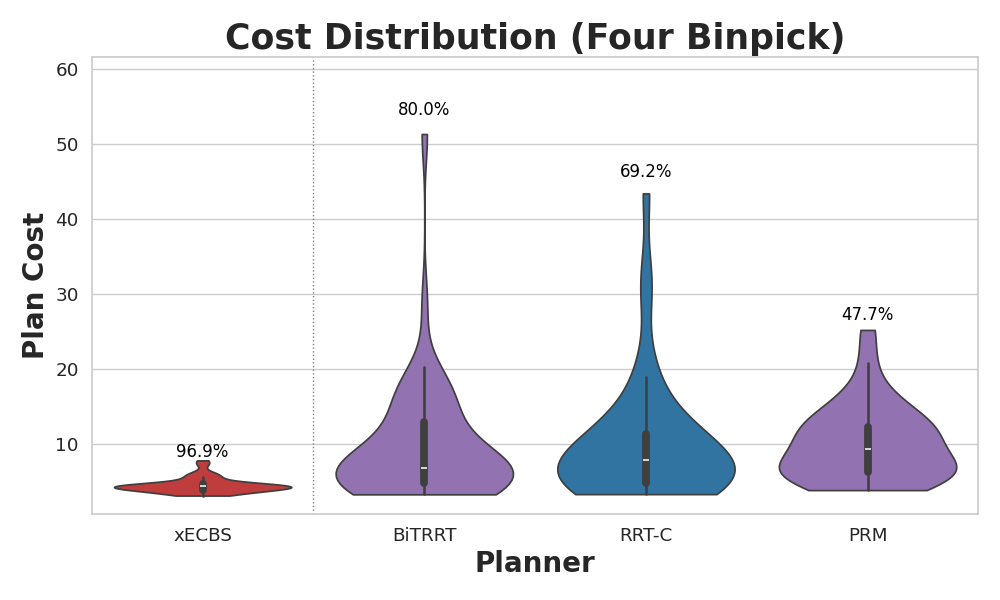}
    \end{subfigure}
    \hfill
    \begin{subfigure}[b]{0.28\textwidth}
        \centering
        \includegraphics[width=\textwidth]{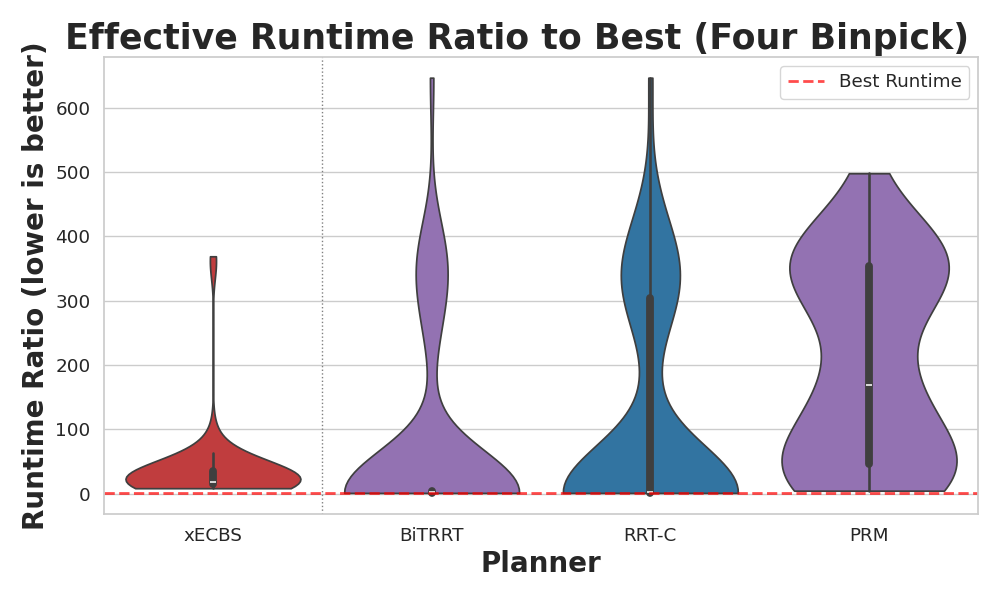}
    \end{subfigure}
    \begin{subfigure}[b]{0.18\textwidth}
        \centering
        \includegraphics[width=\textwidth]{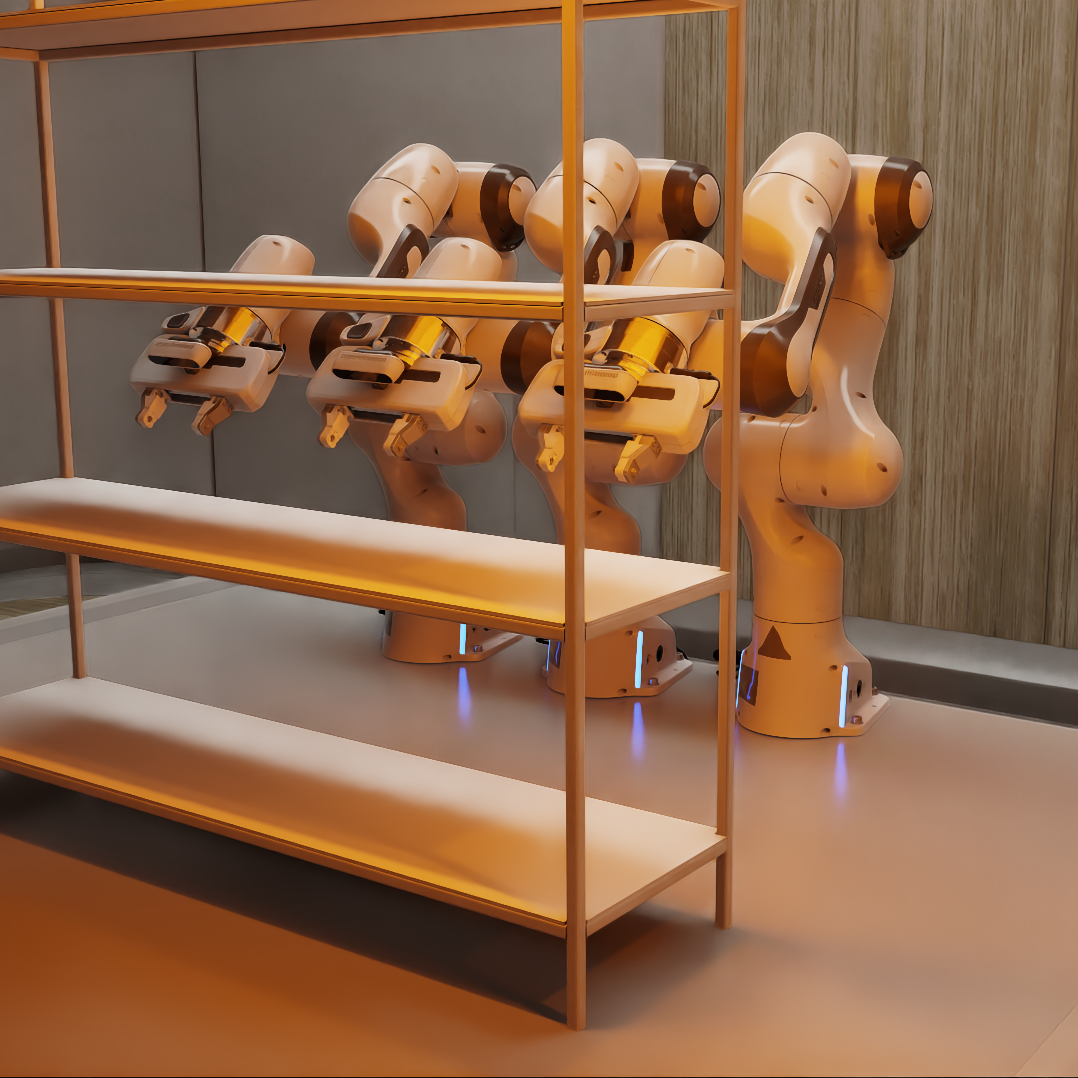}
    \end{subfigure}    
    \centering
    \begin{subfigure}[b]{0.22\textwidth}
        \centering
        \includegraphics[width=\textwidth]{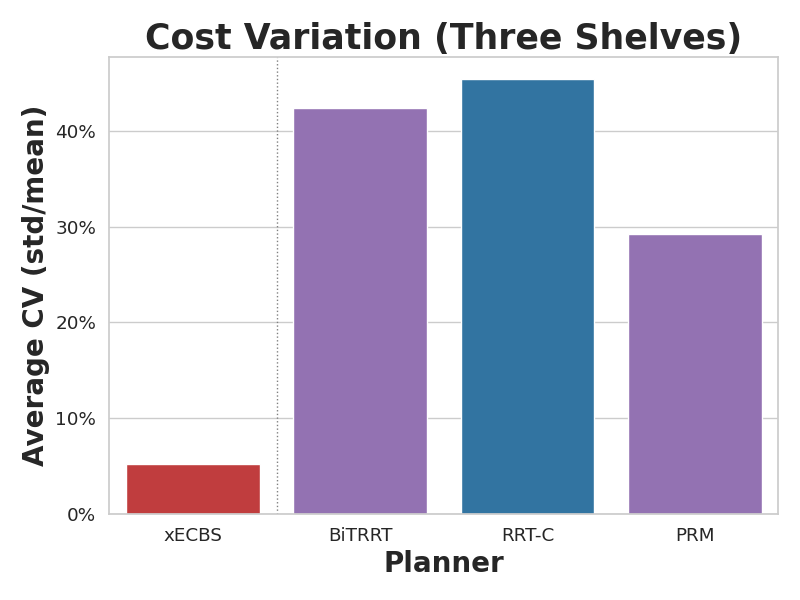}
    \end{subfigure}
    \hfill
    \begin{subfigure}[b]{0.28\textwidth}
        \centering
        \includegraphics[width=\textwidth]{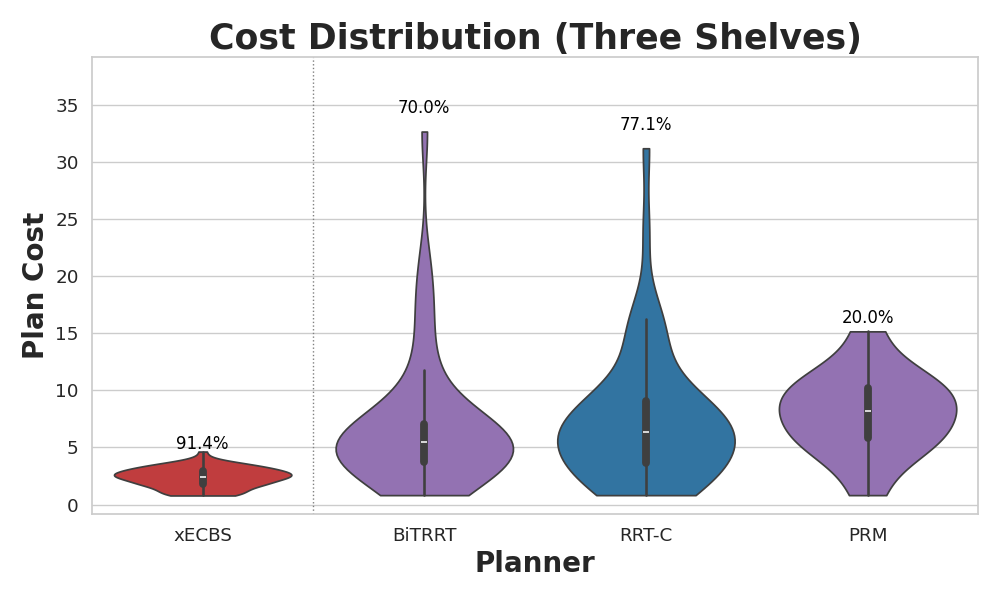}
    \end{subfigure}
    \hfill
    \begin{subfigure}[b]{0.28\textwidth}
        \centering
        \includegraphics[width=\textwidth]{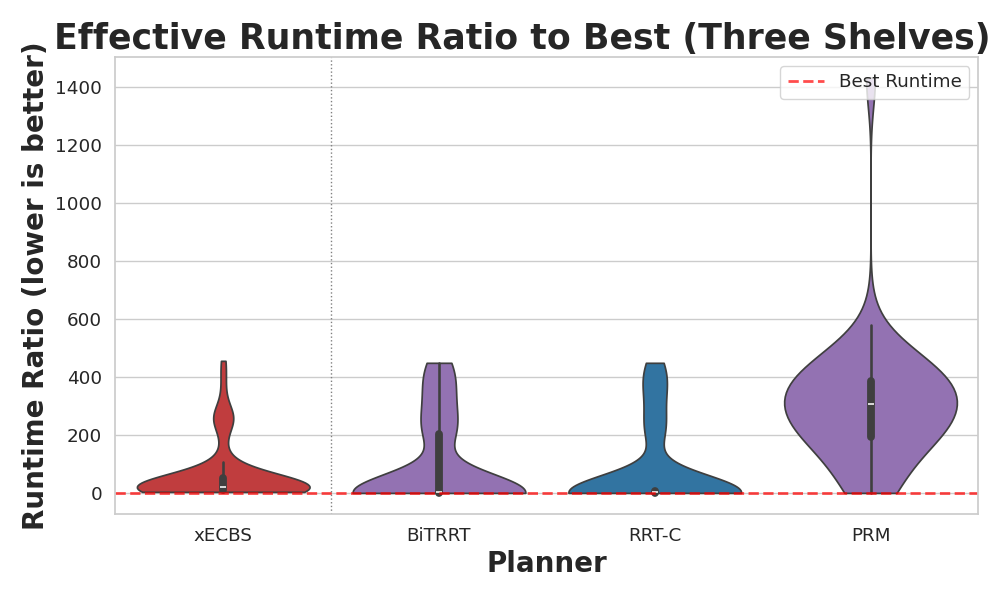}
    \end{subfigure}
    \caption{Multi-arm motion planning evaluations. Top: Four robotic arms plan concurrent motions for picking and placing objects in bins. Bottom: Three robotic arms move between pick-poses on shelves. These problems exhibit a significant degree of interaction between the arms, especially with three arms placed close to each other. For each scenario, algorithms repeated each of 15 planning problems 5 times, each time with a slight variation in their goal configuration.  We compare intra-problem coefficient of variance (CV), overall average solution sum-of-costs (annotated with success rates), and average effective planning time ratios.}
    \label{fig:exps_multi}
\end{figure*}

%% file: figures/fig_benchmark.tex
\begin{figure}[t!]
  \centering
  \begin{subfigure}[b]{\linewidth} 
    \includegraphics[width=\linewidth]{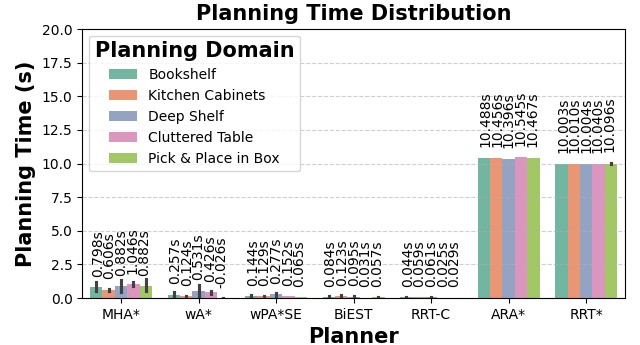}
    \label{fig:mba}
  \end{subfigure}
  \begin{subfigure}[b]{\linewidth} 
    \includegraphics[width=0.97\linewidth]{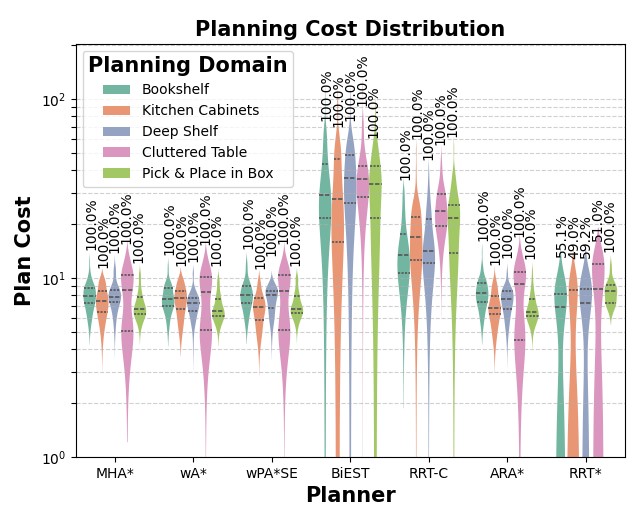}
    \label{fig:mbb}
  \end{subfigure}
    \caption{Evaluating single arm planners of SRMP in a variety of diverse manipulation domains from the Motion Benchmarker \cite{chamzas2022-motion-bench-maker}. We compare the search-based planners of SRMP with the sampling-based planners available in OMPL, namely,  (i) wA* \textit{vs} RRT-C, (ii) MHA* \textit{vs} BiEST and (iii) anytime planners ARA* \textit{vs} RRT*. We also provide results from a parallelized search algorithm (wPA*SE) \cite{pase}. Although OMPL's sampling-based planners have lower planning times in a few instances, their solution paths are long and unpredictable making their planning cost orders of magnitude higher (note the log-scale in the planning cost plot) with large variance. All the planners presented in this evaluation had a maximum planning time of 10s. wPA*SE was allowed to used a maximum of 6 parallel threads.}
  \label{fig:benchmark}
\end{figure}

%% file: sections/background.tex
\section{Background}
\label{sec:background}
Robot motion planning seeks to compute a trajectory that moves a robotic system from a start state to a state satisfying a goal condition. The problem is typically formulated in the robot's configuration space, e.g., the space of possible joint angles in the case of robotic manipulation. Articulated robots with many degrees of freedom (DOF), commonly 6 or 7, require efficient planning methods to explore this space.
In multi-robot manipulation, the dimensionality grows exponentially with the number of robots, making most conventional algorithms impractical for solving the motion planning problem at scale. 

\subsection{Motion Planning Problem}
Let $\mathcal{Q} \subset \mathbb{R}^d$ be the configuration space of a robotic system, with $d$ degrees-of-freedom, operating in a world $\mathcal{W} \subset \mathbb{R}^3$. Additionally, let $\mathcal{O}$ be the set of obstacles in the world. We define the free configuration space $\mathcal{C}_{\text{free}}$ as the set of all configurations in $\mathcal{Q}$ that do not intersect with obstacles in $\mathcal{O}$. Given a start configuration $\conf{}{{\text{start}}} \in \mathcal{C}_{\text{free}}$ and a goal condition $\xi_{\text{goal}}: \mathcal{Q} \rightarrow \{0, 1\}$, the motion planning problem is to find a trajectory $\tau: [0, 1] \rightarrow \mathcal{Q}_{\text{free}}$ such that $\tau(0) = \conf{}{\text{start}}$ and $\xi_{\text{goal}}(\tau(1)) = 1$, that minimizes a cost function. 

When planning for a single robot manipulator, the configuration space $\mathcal{Q}$ is typically defined by the robot's joint angles, 
and the goal condition $\xi_{\text{goal}}$ is a predicate that checks if the end-effector of the robot is within a certain distance of the goal pose, goal position, or that the joint angles are within a certain threshold of a goal configuration.
When planning for multi-robot manipulation, the configuration space $\mathcal{Q}$ is the Cartesian product of the configuration spaces of all robots, i.e., $\mathcal{Q} := \mathcal{Q}^1 \times \mathcal{Q}^2 \times \ldots \times \mathcal{Q}^n$ where $n$ is the number of robots. Here, the goal condition $\xi_{\text{goal}}$ is a predicate that performs similar checks to the single robot case across all $n$ robots. The multi-robot motion planning problem is similar to the single-robot case as well, with the added requirement that robots do not collide with each other. 



\subsection{Algorithms Taxonomy}
Motion planning algorithms are commonly classified into four broad categories: sampling-based, optimization-based, inference-based, and search-based methods. 

\textit{\textbf{Sampling-based methods}}, over the past two decades, have become popular for solving motion planning problems due to their ability to quickly explore high-dimensional configuration spaces and complex environments.
Pioneering algorithms like the Probabilistic Roadmap Method (PRM)~\cite{PRM} and Rapidly-exploring Random Trees (RRT)~\cite{rrt} have been widely adopted and extended \cite{sampling-review,InformedRRT*,RRT*, RRTConnect, bit*} to solve a variety of motion planning problems, and often are the go-to choice for solving motion planning problems for robot manipulators, particularly in research settings.
The core idea behind sampling-based algorithms is to guide the construction of a graph (or a tree) in the configuration space with random continuous samples. 
The random samples can be used directly as vertices in this graph or provide directional cues for graph extension. These algorithms normally seek to find a sequence of consecutive vertices and edges (i.e., a path) in the constructed graphs that connect the start state to a goal state.
Different algorithms use different sampling and extension strategies; the choice of algorithm depends on the problem at hand. 

While random sampling enables efficient motion planning, i.e., finding valid trajectories quickly, it also makes algorithms susceptible to finding very different solutions to nearly identical problems. Furthermore, the inherent randomness in sampling can produce solutions that, while feasible, may appear to humans as ``unintuitive'' or ``sub-optimal'' (see Fig. \ref{fig:exps_single}, left). In certain scenarios, or when seeking short trajectories, the algorithm may require a large number of samples before finding a suitable solution.

These challenges become even more significant in multi-agent settings. The application of the aforementioned algorithms to plan in the high-dimensional composite configuration space $\mathcal{Q}$ -- the configuration space formed by the Cartesian product of all individual configuration spaces -- results in significantly increased computational complexity for finding short paths, making the popular sampling-based approaches less scalable for multi-robot arm motion planning. 

\textit{\textbf{Optimization-based methods}} can generate high-quality trajectories \cite{ratliff2009chomp, kalakrishnan2011stomp, gpmp2, trajopt} that respect robot dynamics.
They formulate the motion planning problem as an optimization problem \cite{optctr} with cost functions defined for trajectory length, time, or energy consumption. After transcribing into a finite-dimensional optimization, these methods rely on the gradients of the cost function and dynamics and employ numerical optimization algorithms to find locally optimal solutions \cite{trajopt, komo, ddp}. However, except for a small subset of systems (such as linear or flat systems), for most nonlinear systems these methods lack guarantees on completeness, optimality or convergence. Consequently, there is growing interest in developing hybrid approaches that combine optimization methods with search-based \cite{insat1, ramthesis} or sampling-based \cite{idba} methods to mitigate these limitations. The prevailing wisdom in optimization-based motion planning is to formulate the problem as a convex optimization to leverage its strong theoretical guarantees and the efficiency of modern solvers. But motion planning for manipulation is fundamentally non-convex due to collision avoidance constraints. Solving these non-convex optimizations requires a well-engineered initial guess which can be highly non-trivial. Recently, Graphs of Convex Sets (GCS) \cite{marcucci2022motion} was proposed for motion planning around obstacles with convex optimization. While promising, these methods require laborious preprocessing to generate the convex sets restricting their use to fixed settings with infrequent changes to the planning scene.

\textit{\textbf{Inference-based methods}} are tightly related to optimization-based methods and use probabilistic models to infer the most likely trajectory given the objective and constraints (and observations) \cite{pmlr-vR4-attias03a, approx-inference, levine2018reinforcement, planning-as-inference, mpd}. These methods are often used in the context of learning from demonstrations, where the goal is to infer the most likely trajectory that a human demonstrator would take to solve a task. While these methods have shown promise in learning from demonstrations, they are often limited by the quality and quantity of the demonstrations available and may not generalize well to unseen scenarios. Collecting human demonstrations at scale is challenging, and in multi-agent settings, this task becomes even more difficult due to the increased complexity of coordination 
\cite{mmd}. As a result, there is a growing demand for motion planners capable of generating motions specifically for data collection purposes. The generated motions must be of high quality and ideally resemble human-like trajectories to improve the generalization of inference-based methods. Our library fits well for those requirements.

\textit{\textbf{Search-based methods}} are a class of algorithms that use a best-first search strategy to explore the configuration space systematically to find a feasible solution. 
These methods have key properties that make them attractive for motion planning problems: they are deterministic, complete, and can often guarantee global optimality or bounded sub-optimality \cite{aine2016mhastar, anytime_dynamicA*, pase, mukherjee2022epase, koenig2004lifelong}.
Although search-based methods are often computationally expensive, they have been adapted to address this limitation and show promise in solving high-dimensional motion planning problems, particularly for manipulation \cite{cohen2010search, cohen2011adaptiveprimitives, egraphs, phillips2015efficient, aine2016mhastar}. These methods rely on three key aspects: they build the graph implicitly (i.e., they do not require the entire graph to be constructed before starting the search but instead build it while searching), they use powerful heuristics to guide the search towards the goal, and they employ motion primitives to generate \textit{lattice} graphs.
Moreover, recent developments have demonstrated that search-based methods can efficiently solve multi-robot manipulation problems \cite{shaoulmishani2024xcbs, shaoul2024gencbs}. These methods have proven capable of scaling to plan motions for dozens of manipulators while providing guarantees on completeness and bounded suboptimality. This makes them particularly well-suited for industrial applications, where reliable and scalable motion planning is crucial, and for data collection pipelines that require structured, predictable motion generation.
However, implementing these methods efficiently is not trivial. \
SRMP provides an interface for search-based motion planning algorithms, enabling researchers and practitioners to easily experiment with and deploy these algorithms for single and multi-robot manipulation problems.

\subsection{Search-Based Planning Algorithms in SRMP}
\input{sections/sbp}

\subsection{Existing Software Packages}
One of the most widely used motion planning libraries is the Open Motion Planning Library (OMPL) \cite{ompl}, an open-source framework that provides implementations of various sampling-based motion planning algorithms. OMPL is designed to be flexible and extensible, allowing users to experiment with different algorithms. It has been integrated into several popular robotics motion planning software packages, including MoveIt! \cite{moveit}, Robowflex \cite{robowflex}, MPLib \cite{mplib}, and AIKIDO, and has been widely adopted in research and development. While OMPL is a powerful tool, it may not always meet the needs of practitioners, particularly in industrial applications and in robot-learning. In both, reliability and predictability of robot motions are critical for safety (e.g., when robots work next to humans) and reproducibility (e.g., for creating consistent training data). In such cases, search-based methods are often preferred due to their deterministic nature and guarantees of completeness and bounded suboptimality. 

Other libraries address different motion planning needs. ROS Motion Planning provides motion planning capabilities for ground robots in the ROS ecosystem but does not support high-dimensional motion planning problems, limiting their applicability to more complex robotic systems. Several libraries focus on optimization-based motion planning algorithms, including TrajOpt \cite{trajopt}, CHOMP \cite{ratliff2009chomp}, and GPMP2 \cite{gpmp2}. These libraries are designed to solve trajectory optimization problems and are often used alongside sampling-based methods to refine solutions. However, integrating them into existing software can be nontrivial and often requires additional effort to enable compatibility.
More recently, CuRobo \cite{curobo} was introduced as a motion planning library that leverages CUDA-based optimization algorithms for efficient and scalable motion planning. While CuRobo offers significant computational advantages, it requires a GPU with CUDA support, which may not be available in all deployment environments. Additionally, integrating CuRobo into existing software stacks can be challenging and may demand substantial development effort. 
Search-Based Planning Library (SBPL) was developed to provide access to search-based motion planning algorithms for single-robot motion planning problems. However, SBPL is no longer actively maintained and does not support modern frameworks. Notably, despite the strengths of these libraries (each for specific cases), none of them provide native support for multi-arm motion planning, which remains a critical gap in existing motion planning frameworks.

%% file: sections/sbp.tex
SRMP provides a variety of search-based algorithms for manipulation motion planning. While each algorithm is different, most are similar in several key characteristics. First, all algorithms use a priority queue to explore the state space of a problem.
The priority queue maintains states that have been generated but not yet expanded. States are expanded based on their priority, determined by a priority function.\footnote{In general, designing effective and admissible priority functions remains a significant challenge.} When this function is consistent or admissible \cite{heuristic-search}, algorithms can guarantee the solution optimality.

Given that the dimensionality of manipulation motion planning problems is typically high, the state space is often prohibitively large, making explicit graph representation impractical. Instead, all algorithms in SRMP also construct an implicit graph using motion primitives defined by the domain's action space. This approach significantly reduces memory requirements, as only the portions of the graph relevant to the current search are stored.

While even simple search-based algorithms like \textbf{wA*} \cite{a*} can achieve impressive performance, SRMP includes improved methods with various strategies. For instance, \textbf{MHA*} \cite{aine2016mhastar}, refines state prioritization by incorporating inadmissible heuristics while maintaining bounded sub-optimality guarantees. Others, like \textbf{ARA*} \cite{anytime_dynamicA*}, focus on quickly finding an initial, potentially suboptimal, solution -- particularly in time-sensitive tasks -- before refining it within a given time limit. Additionally, SRMP includes algorithms that accelerate the search by parallelizing it across multiple cores. These methods, which include \textbf{wPA*SE} \cite{mukherjee2022epase}, are designed to preserve the guarantees of bounded sub-optimality.

While common motion planning algorithms are efficient for single-arm planning, they do not directly scale to multi-robot manipulation due to the exponential growth in state space dimensionality as the number of robots increases. Addressing this, SRMP includes the recently proposed \textbf{xECBS} \cite{shaoulmishani2024xcbs} algorithm that explores this space by decoupling the planning problem into individual arms while still guaranteeing completeness and bounded sub-optimality. This and similar methods find collision-free solutions by planning individually for arms and imposing constraints \cite{shaoulmishani2024xcbs, shaoul2024gencbs}.

%% file: sections/conclusion.tex
\section{Conclusion}
In this paper, we introduced SRMP, a Search-based Robot Motion Planning software designed to reliably produce consistent plans for single- and multi-robot manipulation systems. SRMP is user-friendly, accessible through both Python and C++, and can function as a standalone software or as a MoveIt! plugin. 
SRMP provides motion plans that are reliable, predictable, and consistent. 
Notably, to the best of our knowledge SRMP is the first framework capable of efficiently solving motion planning problems for multi-robot manipulation while offering rigorous guarantees and generating low-cost motions.
We demonstrated SRMP’s effectiveness across a variety of scenarios and provided example code to illustrate its usage. We hope SRMP will be help advance robotics in research and industry.

%% file: sections/acknowledge.tex
\section{Acknowledgments}
We thank the past and present members of the Search-Based Planning Lab at the Robotics Institute, Carnegie Mellon University, for their contributions to advancing robot motion planning through search-based methods. We also acknowledge Hanlan Yang for his work on \textit{wPA*SE}, which contributed to the development of SRMP. 